\documentclass{amia}
\usepackage{graphicx}
\usepackage[labelfont=bf]{caption}
\usepackage[superscript,nomove]{cite}
\usepackage{color}
\usepackage{multirow}
\usepackage{subcaption}
\usepackage{hyperref}
\begin{document}

\title{Generalization of Deep Convolutional Neural Networks - A Case-study on Open-source Chest Radiographs}

\author{Nazanin Mashhaditafreshi, BSc$^{1}$, Amara Tariq, Ph.D.$^{3}$, Judy Wawira Gichoya, MD MS.$^{2,3}$, Imon Banerjee, Ph.D.$^{2, 3}$}

\institutes{
    %$^1$Institution, City, State, Country (if applicable);
    $^1$Department of Computer Engineering, K. N. Toosi University of Technology, Tehran, Iran; 
    $^2$Department of Radiology, Emory School of Medicine, Atlanta, GA, USA;
    $^3$Department of Biomedical Informatics, Emory School of Medicine, Atlanta, GA, USA\\
    }

\maketitle

\noindent{\bf Abstract}

\textit{Deep Convolutional Neural Networks (DCNNs) have attracted extensive attention and been applied in many areas, including medical image analysis and clinical diagnosis. One major challenge is to conceive a DCNN model with remarkable performance on both internal and external data. We demonstrate that DCNNs may not generalize to new data, but increasing the quality and heterogeneity of the training data helps to improve the generalizibility factor. We use InceptionResNetV2 and DenseNet121 architectures to predict the risk of 5 common chest pathologies. The experiments were conducted on three publicly available databases: CheXpert, ChestX-ray14, and MIMIC Chest X-ray JPG. The results show the internal performance of each of the 5 pathologies outperformed external performance on both of the models. Moreover, our strategy of exposing the models to a mix of different datasets during the training phase helps to improve model performance on the external dataset.
}

\section*{Introduction}
The proliferation of big data coupled with non-linear data abstraction (filters) and high performance computing~\cite{van2020beyond} has spurred rapid advancement in deep learning applications including speech recognition, sentiment analysis, computer vision, and machine translation. These areas were previously thought to be extremely hard for computers to analyze and required hundreds of hours of manual feature engineering yet deep learning techniques deliver state-of-the-art performance with minimal human intervention. Medicine is witnessing rapid adoption and application of deep learning. For example, a large volume of radiology studies are performed daily in most centers yet the number of available trained radiologists remains constant~\cite{hosny2018artificial}. The opportunity to standardize the clinical workflow is thus seen as a low hanging fruit for automation using deep learning, with lots of efforts deemed as hype that try to replace radiologists using deep learning.

Deep Convolutional Neural Networks (DCNNs) apply multiple layers of convolution operations to extract translation and scale invariant features from images, and are widely used to analyze radiology image content to assist in diagnosis. DCNNs have achieved expert-level performance for various chest pathologies \cite{CheXNet,Majkowska2020}. Beyond classification tasks on radiology images, researchers have attempted to rebuild the imaging workflow, assessing DCNNs performance after non-image data fusion for classification of multi-label chest X-ray images \cite{Baltruschat2019}. Despite a plethora of multiple publications improving on the state-of-the-art, validation and scalability of deep learning in medicine remains limited, since model development and validation is frequently performed on a single institutional dataset. A review of studies published in 2018 found that only  6\% (31 of 516) of published studies performed external validation (i.e. studies had a diagnostic cohort design, included data from multiple institutions, and performed prospective data collection).\cite{kim2019design}

Overfitting is a well-known limitation of complex DCNN models which may produce an overly optimistic performance. Therefore, it is important for an optimized DCNN model to have sustained performance on unseen external datasets to promote model generalizibility and translation of models to real life clinical work. Despite the expensive cost of labeling medical datasets, there are several publicly available datasets for chest radiographs that can be used for testing the model generalization. These datasets include the MIMIC Chest X-ray JPG (MIMIC-CXR-JPG) Database v2.0.0 from Beth Israel Deaconess Medical Center in Boston \cite{johnson_peng_lu_mark_berkowitz_horng_2019,DBLP:journals/corr/abs-1901-07042,Goldberger2000}; the CheXpert dataset released from the Stanford Hospital, performed between October 2002 and July 2017 coded with 14 common radiographic diseases \cite{DBLP:journals/corr/abs-1901-07031}; and the ChestX-ray14 dataset from the U.S. National Institutes of Health \cite{DBLP:journals/corr/WangPLLBS17}. The labels for these three datasets were derived from radiology text reports using natural language processing algorithms. A large number of publications have been published from these three datasets focusing on novel DCNN design and development, and present state-of-the-art performance. However, there are only a limited number of studies on the generalizability of a DCNN model trained on chest X-ray images, specifically assessing whether the model retains its performance and generalizes well on unseen datasets \cite{Zech2018}.

In this study, we perform thorough experiments to understand the generalizability of state-of-the-art DCNN models using data from three publicly available chest radiograph datasets. We selected 5 common pathologies (Cardiomegaly, Edema, Atelectasis, Consolidation, Pleural Effusion) and trained two state-of-the-art DCNN models (DenseNet121 \cite{DBLP:journals/corr/HuangLW16a}, InceptionResNetV2 \cite{DBLP:journals/corr/SzegedyIV16}). To evaluate the performance, we adopted performance metrics that have been previously published for disease recognition tasks \cite{DBLP:journals/corr/abs-1810-00736}. We compared the external and internal performance of the models by training them on different partitions of data from the three datasets, and subsequently tested each model with various combination of test sets of these datasets. We report the test AUC of each experiment which shows that the performance of the DCNNs on internal data outperforms the external performance for the test sets.

\section*{Materials and Methods}
In this section, we present details of the public datasets and architecture of the two DCNN models that were used for our experiments. We also describe the experimentation details to test the generalization capability of the DCNN models. 

\subsection*{Datasets}
 
The CheXpert \cite{DBLP:journals/corr/abs-1901-07031} dataset comprises of $224,316$ frontal and lateral chest radiographs of $65,240$ patients. The ChestX-ray14 \cite{DBLP:journals/corr/WangPLLBS17} dataset contains $112,120$ frontal-view X-ray images of $30,805$ patients. The MIMIC-CXR-JPG \cite{johnson_peng_lu_mark_berkowitz_horng_2019,DBLP:journals/corr/abs-1901-07042, Goldberger2000} dataset consists of $377,110$ images of $65,379$ patients. We selected $5$ diseases (Atelectasis, Edema, Pleural Effusion, Consolidation, Cardiomegaly) which were common among the above datasets. We randomly split the CheXpert dataset into training ($161,035$ images), validation ($18,097$ images), and test ($44,282$ images) sets. ChestX-ray14 dataset was also randomly divided into training ($80,657$ images), validation ($9,169$ images), and test ($22,294$ images) sets. There were no overlapping patients between the training, validation and test sets for the CheXpert and ChestX-ray14 datasets. In addition, we preserved the original test set of the MIMIC-CXR-JPG dataset ($5,159$ images) and considered it as an external test set for every configuration.

\begin{figure}[H]
\begin{subfigure}{0.48\textwidth}
\centering
\includegraphics[width=\linewidth]{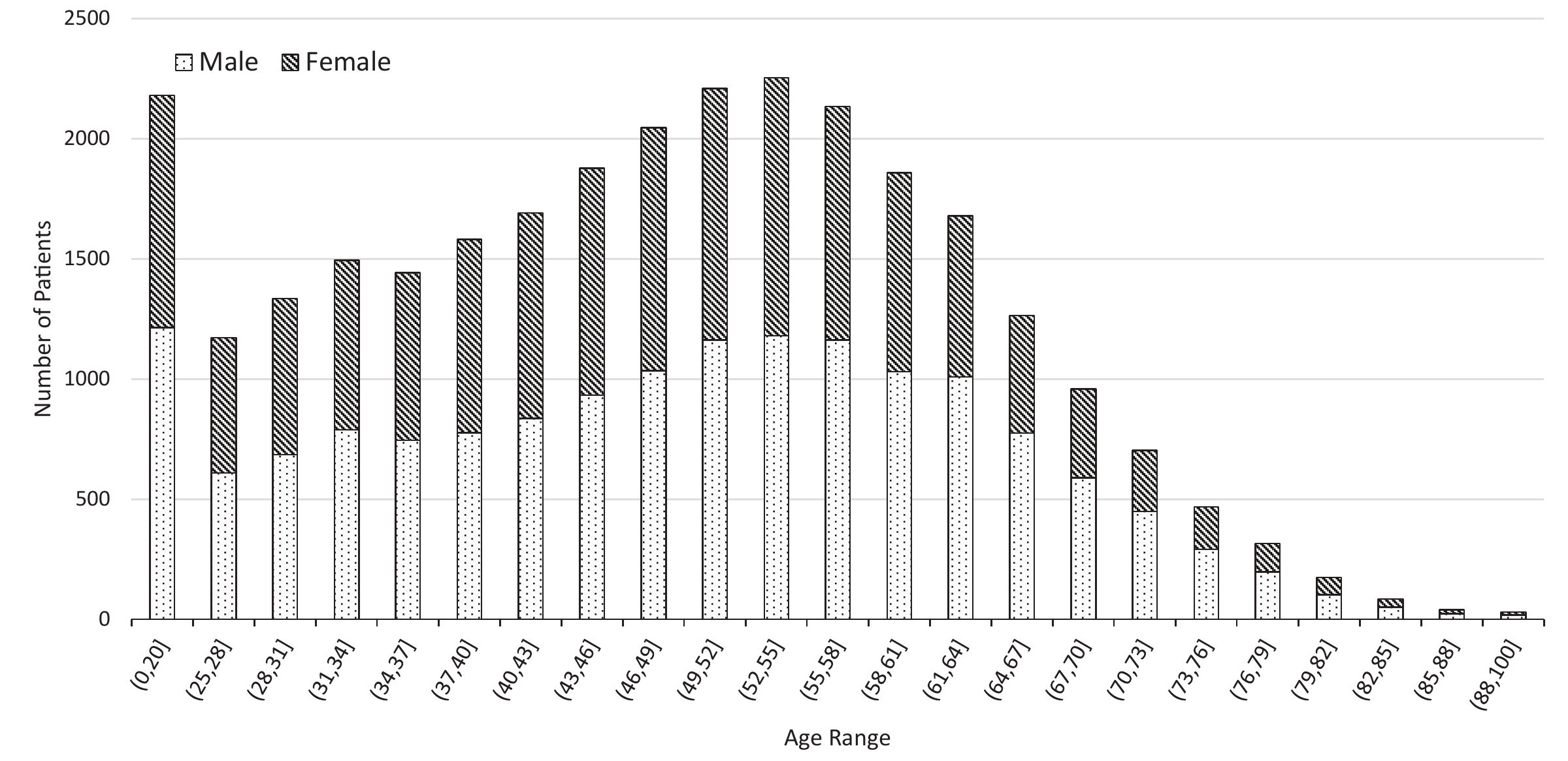}
\caption{ChestX-ray14 Dataset}
\label{nih_age_sex_fig}
\end{subfigure}
\begin{subfigure}{0.48\textwidth}
\centering
\includegraphics[width=\linewidth]{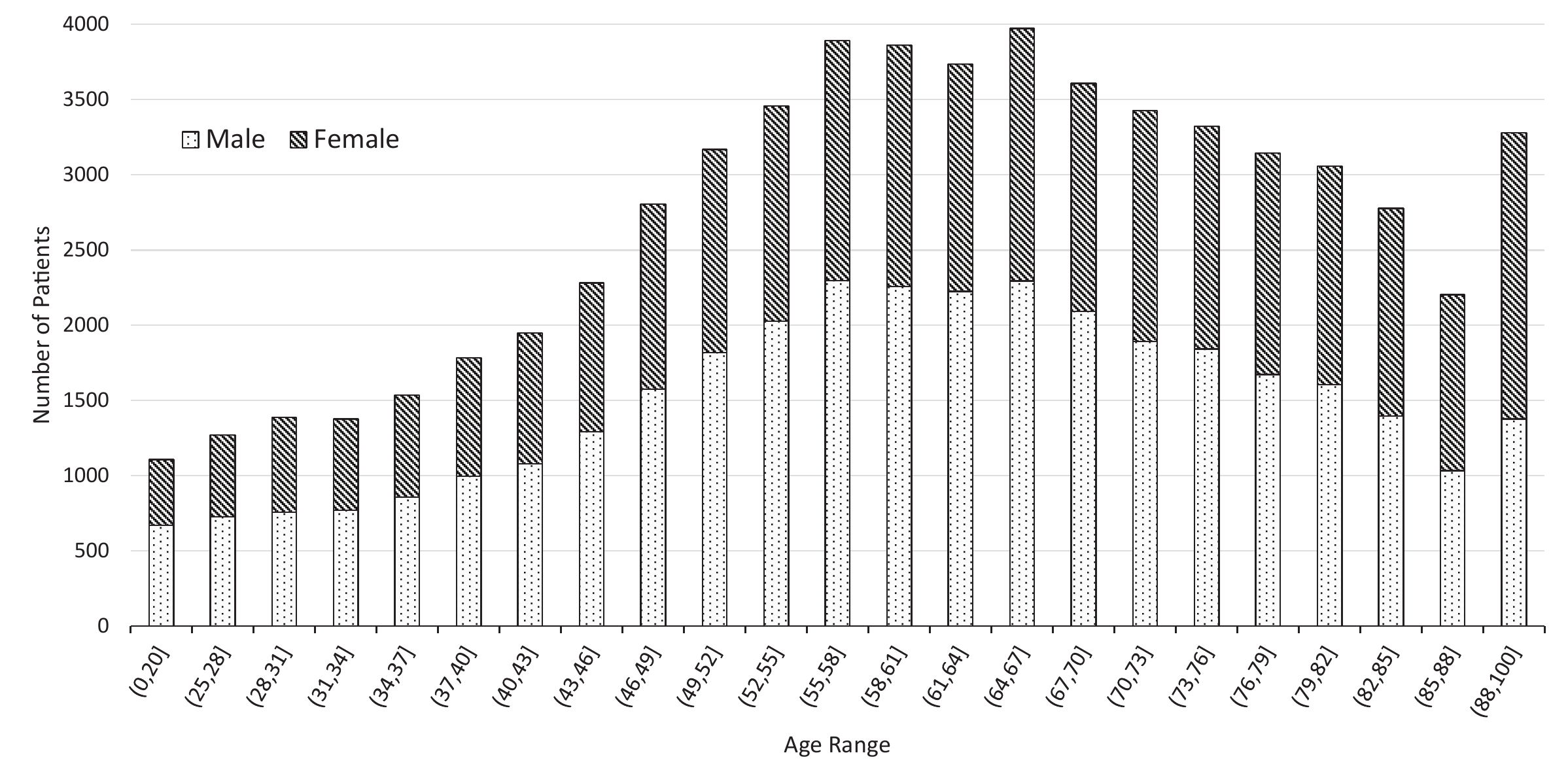}
\caption{CheXpert Dataset}
\label{chex_age_sex_fig}
\end{subfigure}
\caption{Distributions of patients' age and gender}
\end{figure}

The ChestX-ray14 and CheXpert datasets provide some non-image features such as age, gender, and radiographic positioning. Figure \ref{nih_age_sex_fig} shows the distribution of patients' age and gender for the ChestX-ray14 dataset. The average age is $46.13$ years with a standard deviation of $17.01$ years for this dataset. Patients' age and gender distributions for CheXpert dataset are shown in Figure \ref{chex_age_sex_fig}. The average age and the standard deviation are $60.31$ and $18.56$ years, respectively. The distribution of gender is quite similar among the two datasets, but the ChestX-ray14 dataset has a larger proportion of younger patients as compared to the CheXpert dataset.

\subsection*{Data Preprocessing}
For the ground truth labels, we used the binary mapping approach for handling uncertainty in labels, in which the uncertain labels were replaced with 1 (U-Ones model), or 0 (U-Zeroes model). We used U-Ones model for uncertain labels for Atelectasis, Edema, and Pleural Effusion and U-Zeroes model for Cardiomegaly and Consolidation based on the CheXpert results \cite{DBLP:journals/corr/abs-1901-07031}.

For image preprocessing, we applied contrast-limited adaptive histogram equalization (CLAHE) technique~\cite{pizer1987adaptive} for contrast enhancement on all training and validation images before feeding them into the network. Thereafter, we normalized all images based on the mean and standard deviation of images in the ImageNet~\cite{deng2009imagenet} training set. All the images were resized to $299 \times 299$ pixels for InceptionResNetV2 and $224 \times 224$ pixels for DenseNet121 architecture. The scikit-image transform module, which applies first order spline interpolation for image downscaling and Gaussian filter to eliminate aliasing artifacts, was used to resize the images. A constant value of 0 was used to fill the points outside the input boundaries. 50\% of the training data was augmented with random horizontal flipping. 

\subsection*{Model Architecture and Implementation}
We compared two well-known architectures; $i)$ DenseNet121 \cite{DBLP:journals/corr/HuangLW16a}, and $ii)$ InceptionResNetV2 \cite{DBLP:journals/corr/SzegedyIV16}. InceptionResNetV2 combines the Inception architecture (which is a very deep convolutional neural network) with residual connections while DenseNets are used to simplify the connectivity pattern between layers by connecting all layers directly with each other. While residual connections (used in Inception networks) sum up outputs of multiple layers, DenseNets concatenate outputs of multiple connected layers. DenseNets are known to avoid learning redundant feature maps and have much better feature reuse than traditional convolutional neural networks. 

\begin{figure}[h!]
\centering
\includegraphics[width=0.75\linewidth]{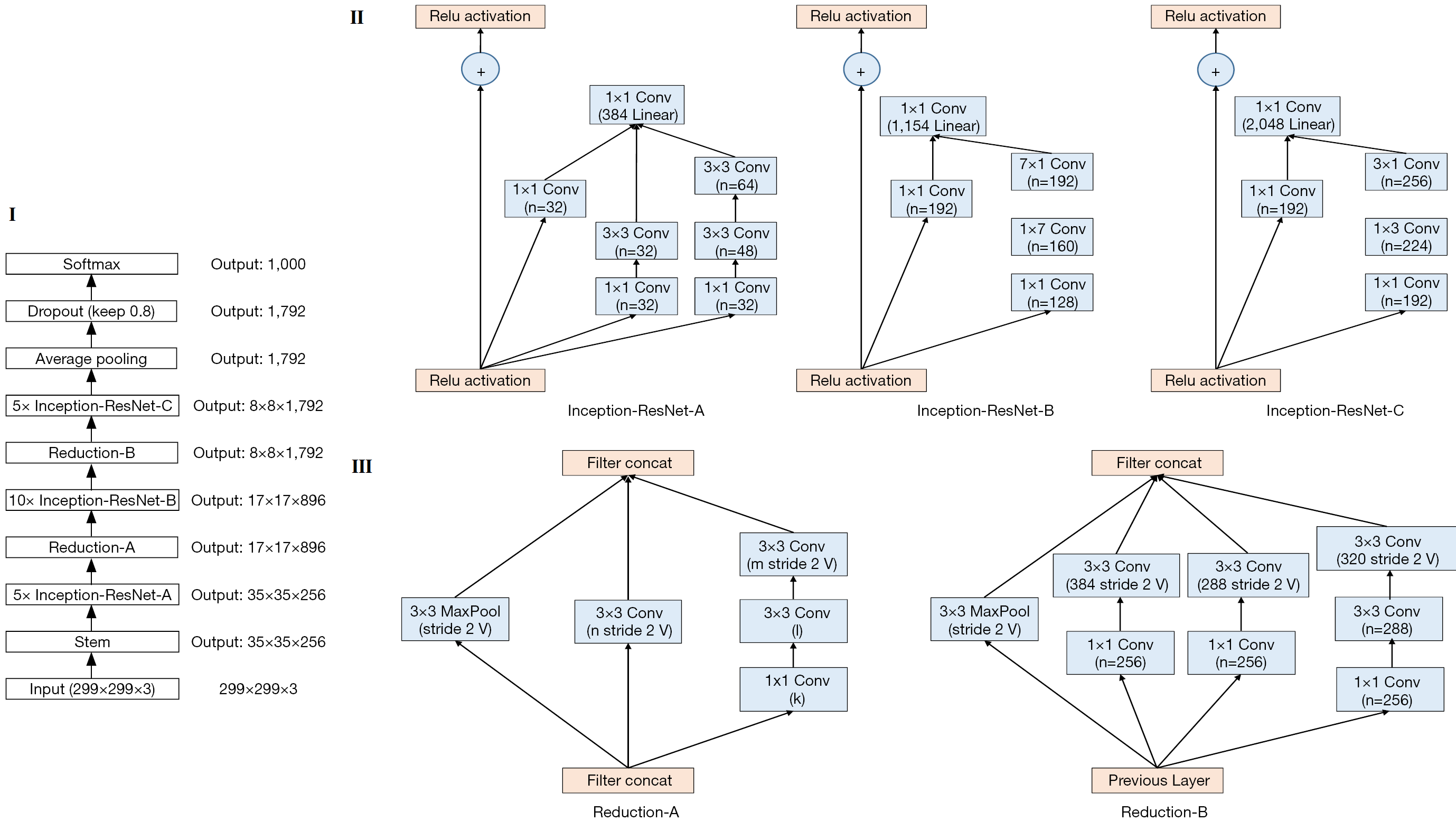}
\caption{InceptionResNetV2 Architecture. ($I$) Schematic of the InceptionResNetV2 model; ($II$) A, B, C are Inception modules which comprise of several convolutional layers; ($III$) A and B are reduction modules which reduce the size of the output. \cite{ATM28771}}
\label{InceptionArchitecture}
\end{figure}

\begin{figure}[h!]
\centering
\includegraphics[width=0.75\linewidth]{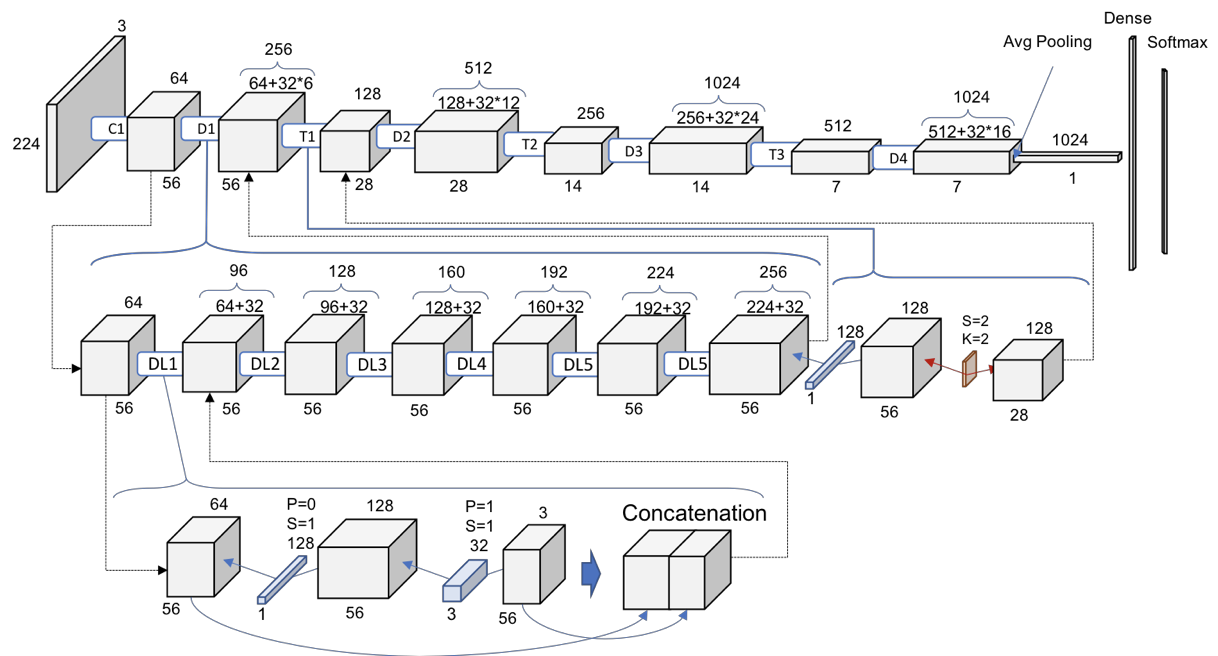}
\caption{DenseNet121 Architecture (Dx: Dense Block x, Tx: Transition Block x, DLx: Dense Layer x) \cite{ruiz_2018}}
\label{DenseNetArchitecture}
\end{figure}

InceptionResNetV2 consists of $782$ layers and $54,283,877$ trainable parameters (Figure \ref{InceptionArchitecture}) and is more complex than DenseNet121 which has $429$ layers and $6,958,981$ trainable parameters (Figure \ref{DenseNetArchitecture}). It intuitively follows that DenseNet121 requires less memory than InceptionResNetV2 and is less susceptible to the vanishing-gradient problem. InceptionResNetV2 achieves better top-1 accuracy on ImageNet-1k validation set for image classification task \cite{DBLP:journals/corr/abs-1810-00736}.


We selected these two architectures to assess the impact of model complexity on external test performance. We largely preserved each architecture while adapting it to our classification task. We removed the top layer and replaced it with a global average 2D pooling layer and a dense layer with sigmoid activation as the last fully-connected layer. Parameter $n$ was set to $5$ for the last layer to match the $5$ labels of our classification problem.  

%Amara: please confirm if the last layer was 'added' or the last layer of the original architecture was 'replaced' with n=5 layer.

We used  binary cross-entropy loss and Adam accumulate optimizer for training each network starting with pre-trained ImageNet weights. For model training, we set  $learning\; rate =1 \times 10^{-3}, decay = 1\times 10^{-5}$, and $batch\;size = 32$. We used GTX 1080 Ti GPU and Keras Python library.

%IMON: NEED SOME MORE INFORMATION ABOUT THE ARCHITECTURE: 1) WHY YOU SELECTED THESE TWO ARCHITECTURES ONLY? 2) HOW THESE ARCHITECTURES ARE DIFFERENT? 3) EXPLAIN BRIEFLY THE LAYERS.
\subsection*{Experiments}
\label{sc:experiments}
\begin{figure}[H]
\centering
\includegraphics[width=0.73\linewidth]{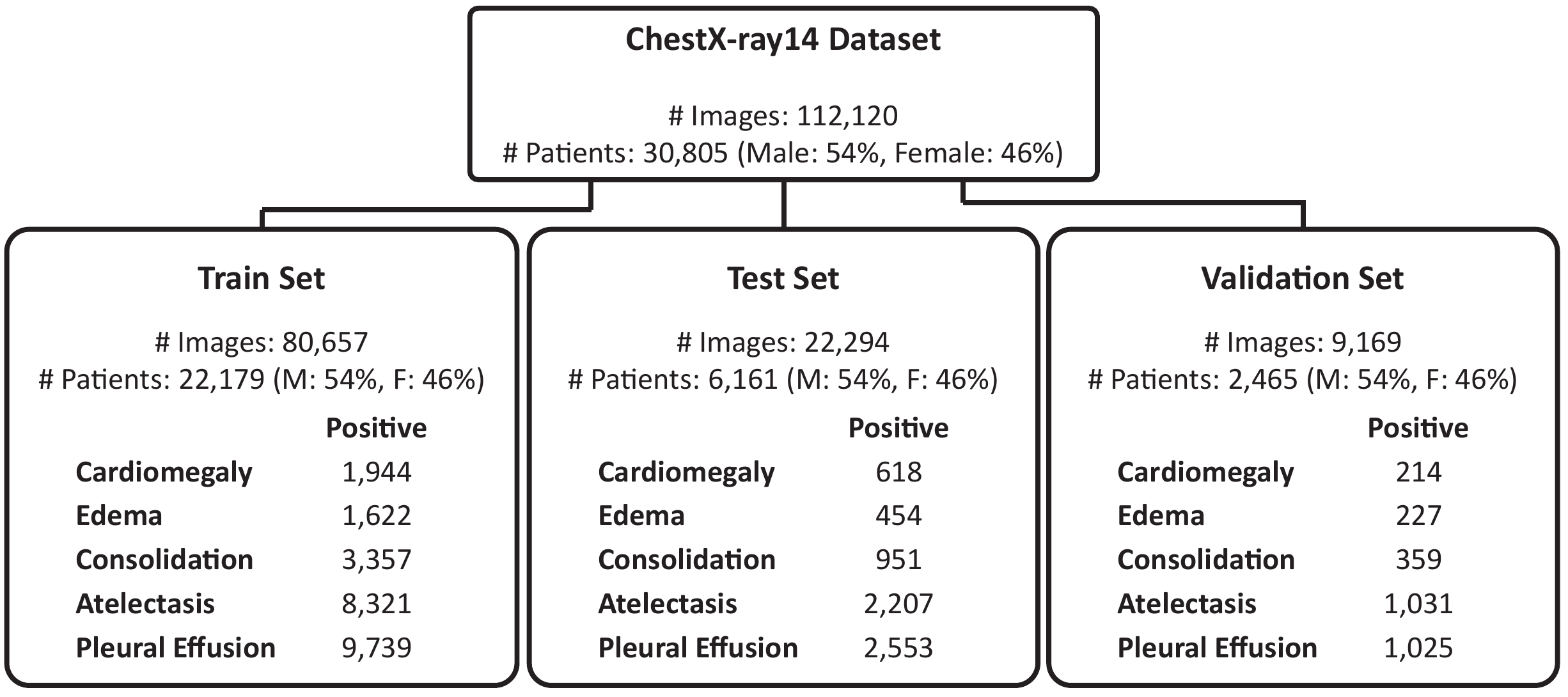}
\caption{Flowchart of images used in this study from the ChestX-ray14 dataset for train, validation, and test sets.}
\label{NIH_fig}
\end{figure}

\begin{figure}[H]
\centering
\includegraphics[width=0.73\linewidth]{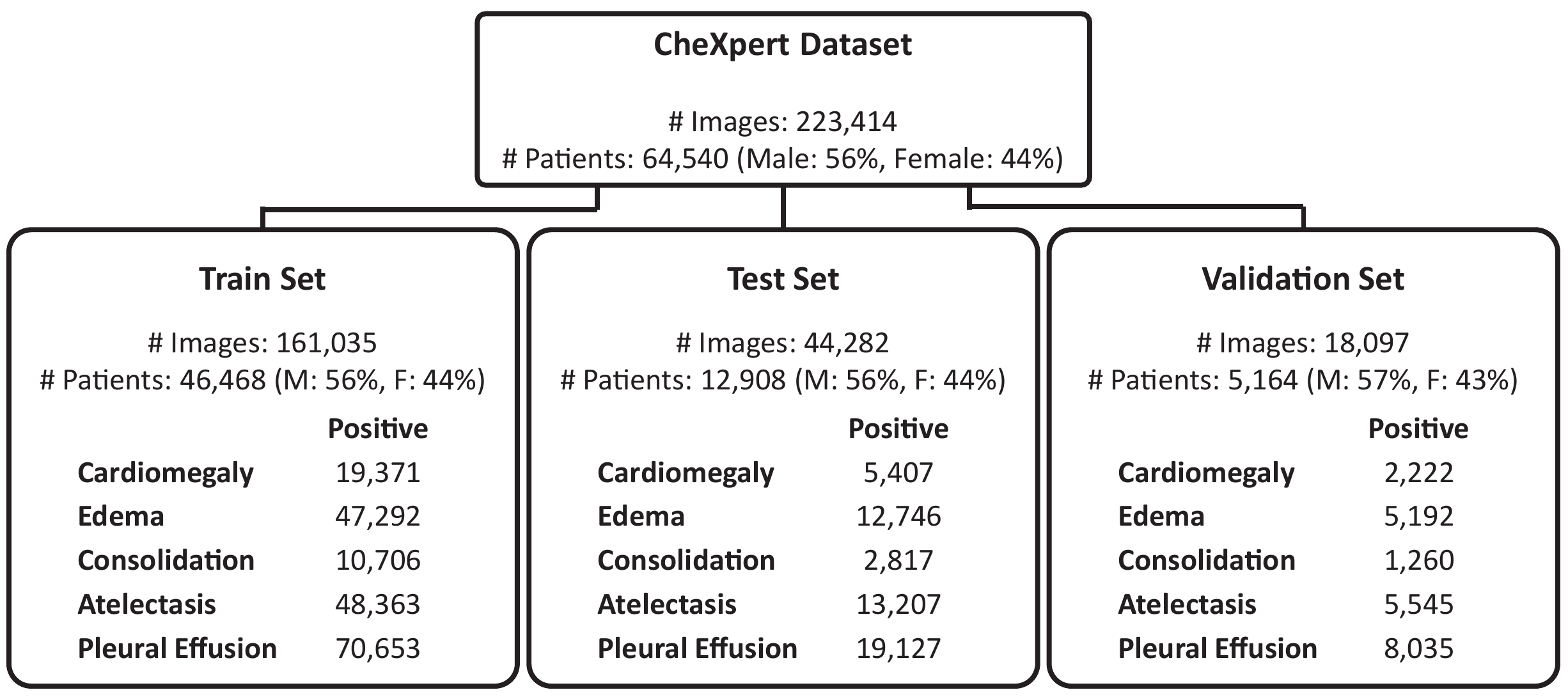}
\caption{Flowchart of images used in this study from the CheXpert dataset for train, validation, and test sets.}
\label{CheX_fig}
\end{figure}

\begin{figure}[H]
\centering
\includegraphics[width=0.73\linewidth]{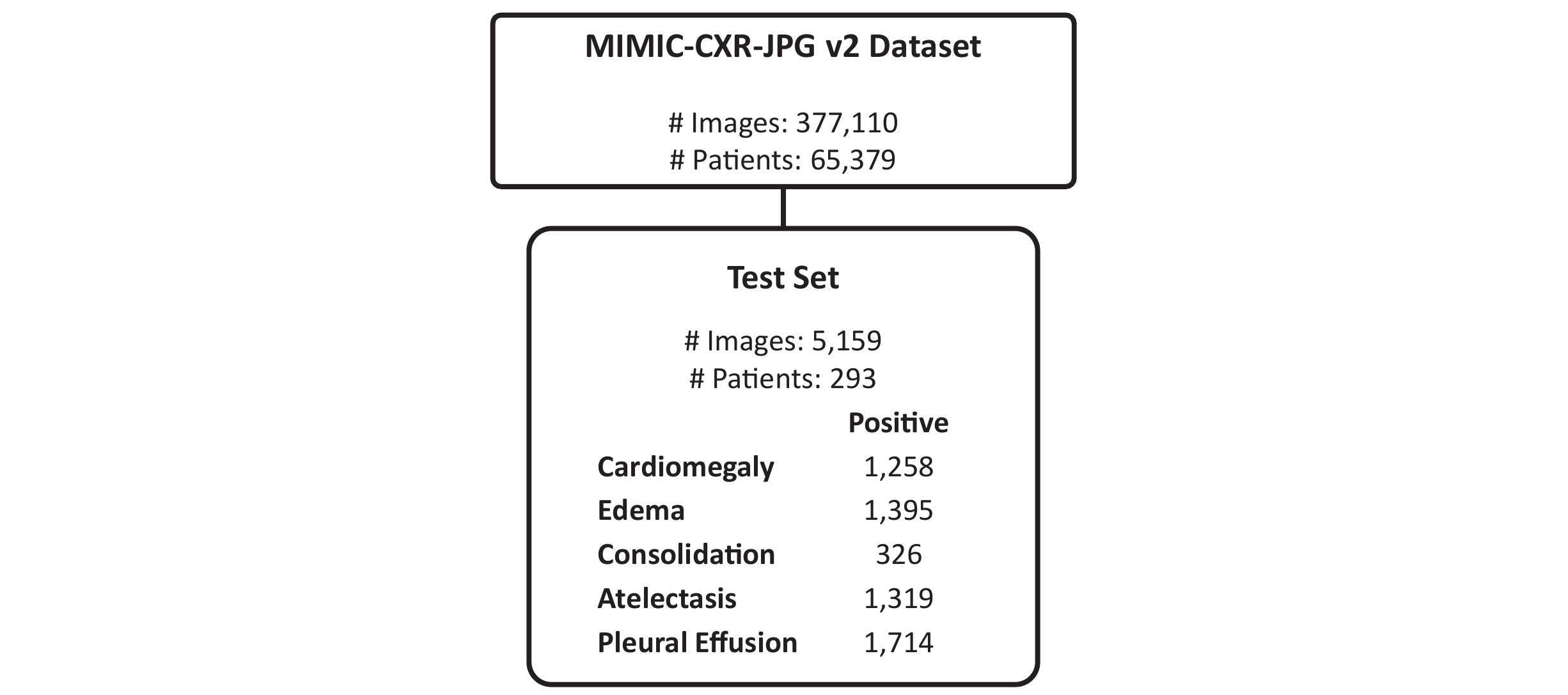}
\caption{Flowchart of images used in this study from the MIMIC-CXR-JPG dataset for test set.}
\label{MIMIC_fig}
\end{figure}

We used three distinct ways to assess the generalization capabilities of state-of-the-art neural networks. For ChestX-ray14 and CheXpert, we randomly split the patients to train, validation, and test sets (Figure \ref{NIH_fig}, \ref{CheX_fig}). For the MIMIC-CXR-JPG, we kept the original test set of $5,159$ images from $293$ patients (Figure \ref{MIMIC_fig}). The details of the three evaluation configurations are described in detail below.

$1)$ In the first configuration, we trained our models on the train set of CheXpert. We tested the models on the test set of the CheXpert dataset for the in-sample data and the test set of the MIMIC-CXR-JPG dataset as the external test set. $2)$ In the second configuration, we trained our models on the train set of ChestX-ray14. Models were tested on the test set of the ChestX-ray14 dataset as the in-sample test and the test set of the MIMIC-CXR-JPG dataset as the external test. $3)$ In the third configuration, to increase the variation of training samples, we trained our models on the combination of train sets of CheXpert and ChestX-ray14 datasets. Models were tested on combined test sets of CheXpert and ChestX-ray14 datasets as the in-sample and the MIMIC-CXR-JPG test set as the external test.

\section*{Results}
%NEED A FIGURE (SCATTER PLOT) COMPARING PERFORMANCE
In this section, we present a discussion on the results of our experiments. We also include observations made by a radiologist on the performance of the two models.

\subsection*{Model Performance}
We performed thorough experimentation to evaluate the robustness and generalization capabilities of two very popular deep learning classification architectures, i.e., InceptionResNetV2 and DenseNet121, for the classification task of common chest diseases diagnosis. As explained in the previous section, we used three different training schemes and tested each trained model over internal and external test sets. Our diagnosis task includes $5$ labels (Cardiomegaly, Edema, Atelectasis, Consolidation, Pleural Effusion) common to the three datasets that we used. We employed AUC (area under the curve of receiver operating characteristics curve) as an evaluation measure. The three public datasets have severe class imbalance for the five selected pathologies. The overall performance on both models in terms of AUC values is reported in Table \ref{tb:test-aucs}. % We have imbalanced classes for each of the $5$ pathologies. Thus, we used AUC metric to measure the ranking quality of each classifier. AUC values are reported in Table \ref{tb:test-aucs}. The following are our observations. 

The two architectures have widely different complexity in terms of the depth of the network as well as the number of trainable parameters. Still, both architectures have the same performance on internal and external test sets. In general, the performance of each model on the internal test set is better than its performance on the external test set for every configuration. As demonstrated in Table \ref{tb:test-aucs}, both architectures have better performance for the diagnosis of Cardiomegaly and Edema on the internal test set compared to the external test set under evaluation configuration $1$ (training over CheXpert dataset). On the other hand, performance for diagnosis of Atelectasis, Consolidation, and Pleural Effusion is quite similar for internal and external test sets under the same configuration for both architectures.

Both architectures have much better performance on internal test sets for all labels under configuration $2$ (training over ChestX-ray14 dataset) as compared to corresponding performance values under configuration $1$. On the other hand, performance on external test sets is worse than the corresponding internal test performance for both architectures for all labels under this configuration.

\renewcommand{\arraystretch}{2}
\begin{table}[H]
\centering
\medskip
\caption{Performance of InceptionResNetV2 and DenseNet121 models for all evaluation configurations for internal and external test sets in terms of AUC-ROC value}
%{AUC of Cardiomegaly, Edema, Consolidation, Atelectasis, and Pleural Effusion for different combinations of train, validation, and test sets of Chest X-ray14, CheXpert, and MIMIC-CXR data sets.}
\vspace{-2mm}
\begin{tabular}{|l|l|l|l|l|l|l|l|l|}
 \hline 
  \rotatebox{90}{\textbf{Model}} &
  \rotatebox{90}{\parbox{4cm}{\textbf{Evaluation Configuration} \\ \textbf{(Train/Validation set)}}} &
  \rotatebox{90}{\textbf{Comparison Type}} &
  \rotatebox{90}{\textbf{Test Set} } &
  \rotatebox{90}{\textbf{Cardiomegaly}} &
  \rotatebox{90}{\textbf{Edema}} &
  \rotatebox{90}{\textbf{Consolidation}} &
  \rotatebox{90}{\textbf{Atelectasis}} &
  \rotatebox{90}{\textbf{Pleural Effusion}} \\ \hline 
  \multirow{6}{*}{\rotatebox{90}{DenseNet121}} &
  
  \multirow{2}{*}{\shortstack[l]{Configuration $1$ \\(CheXpert)}} &
  Internal &
  CheXpert  &
  $0.853$ &
  $0.852$ &
  $0.714$ &
  $0.711$ &
  $0.864$ \\ \cline{3-9} 
 &
   &
  External &
  MIMIC-CXR-JPG  &
  $0.735$ &
  $0.808$ &
  $0.706$ &
  $0.704$ &
  $0.854$ \\ \cline{2-9} 
 &
  \multirow{2}{*}{\shortstack[l]{Configuration $2$ \\(ChestX-ray14)}} &
  Internal &
  ChestX-ray14  &
  $0.905$ &
  $0.88$ &
  $0.797$ &
  $0.803$ &
  $0.873$ \\ \cline{3-9} 
 &
   &
  External &
  MIMIC-CXR-JPG  &
  $0.674$ &
  $0.732$ &
  $0.614$ &
  $0.679$ &
  $0.792$ \\ \cline{2-9} 
 &
  \multirow{2}{*}{\shortstack[l]{Configuration $3$ \\(CheXpert and \\ ChestX-ray14)}} &
  Internal &
  CheXpert and ChestX-ray14  &
  $0.878$ &
  $0.900$ &
  $0.747$ &
  $0.775$ &
  $0.893$ \\ \cline{3-9} 
 &
   &
  External &
  MIMIC-CXR-JPG  &
  $0.744$ &
  $0.817$ &
  $0.694$ &
  $0.729$ &
  $0.859$ \\ \hline
\multirow{6}{*}{\rotatebox{90}{InceptionResNetV2}} &
  \multirow{2}{*}{\shortstack[l]{Configuration $1$ \\(CheXpert)}} &
  Internal &
  CheXpert &
  $0.854$ &
  $0.857$ &
  $0.719$ &
  $0.716$ &
  $0.866$ \\ \cline{3-9} 
 &
   &
  External &
  MIMIC-CXR-JPG  &
  $0.729$ &
  $0.815$ &
  $0.694$ &
  $0.716$ &
  $0.856$ \\ \cline{2-9} 
 &
  \multirow{2}{*}{\shortstack[l]{Configuration $2$ \\(ChestX-ray14)}} &
  Internal &
  ChestX-ray14 &
  $0.910$ &
  $0.871$ &
  $0.796$ &
  $0.805$ &
  $0.870$ \\ \cline{3-9} 
 &
   &
  External &
  MIMIC-CXR-JPG &
  $0.651$ &
  $0.702$ &
  $0.626$ &
  $0.682$ &
  $0.773$ \\ \cline{2-9} 
 &
  \multirow{2}{*}{\shortstack[l]{Configuration $3$ \\(CheXpert and \\ ChestX-ray14)}}  &
  Internal &
  CheXpert and ChestX-ray14 &
  $0.881$ &
  $0.902$ &
  $0.752$ &
  $0.781$ &
  $0.894$ \\ \cline{3-9} 
 &
   &
  External &
  MIMIC-CXR-JPG  &
  $0.734$ &
  $0.808$ &
  $0.706$ &
  $0.724$ &
  $0.862$ \\ \hline
\end{tabular}
\label{tb:test-aucs}
\end{table}
\vspace{-5mm}
Configuration $3$ involves training over a larger and more generalized set by combining training sets of ChestX-ray14 and CheXpert datasets. We observed similar performance comparisons between internal and external test sets for both architectures as was observed in the other two configurations. Performance for internal sets is generally better than performance for external sets. There is a mixed trend in terms of performance improvement for internal sets as compared to corresponding performance values for the other two configurations. For both architectures, AUC for Edema and Pleural Effusion are better than corresponding values of other configurations for the same models. On the other hand, external test set performance improves for almost all labels under this configuration for both models. Hence, generalized training sets seem critical in improving the generalization capabilities of trained models.

\begin{figure}[H]

\begin{subfigure}{0.32\textwidth}
\centering
\includegraphics[width=\linewidth]{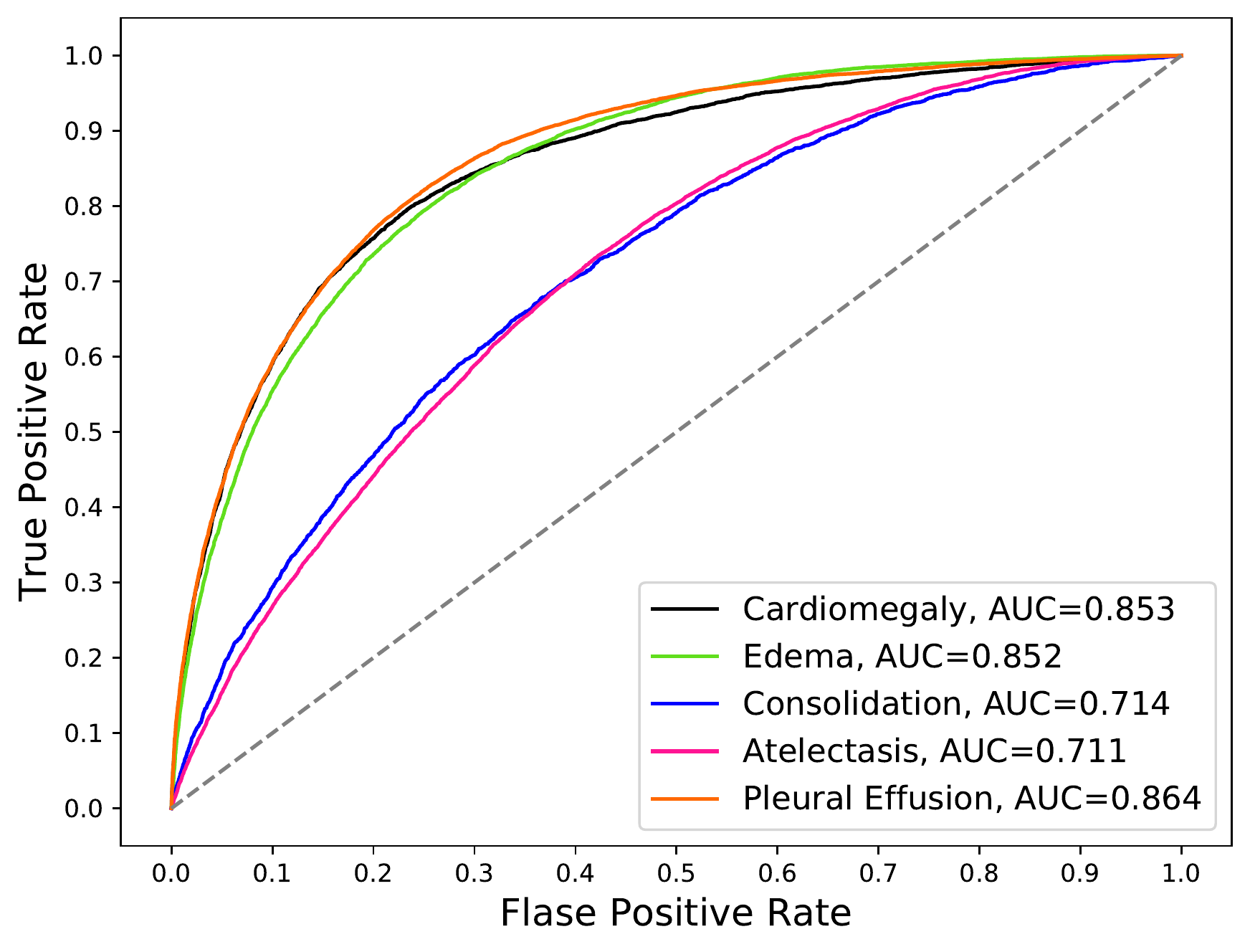}
\centering 
\captionsetup{justification=centering}
\caption{Train: CheXpert \\ Test: Internal}
\label{DenseNet1_chex}
\end{subfigure}
\begin{subfigure}{0.32\textwidth}
\centering
\includegraphics[width=\linewidth]{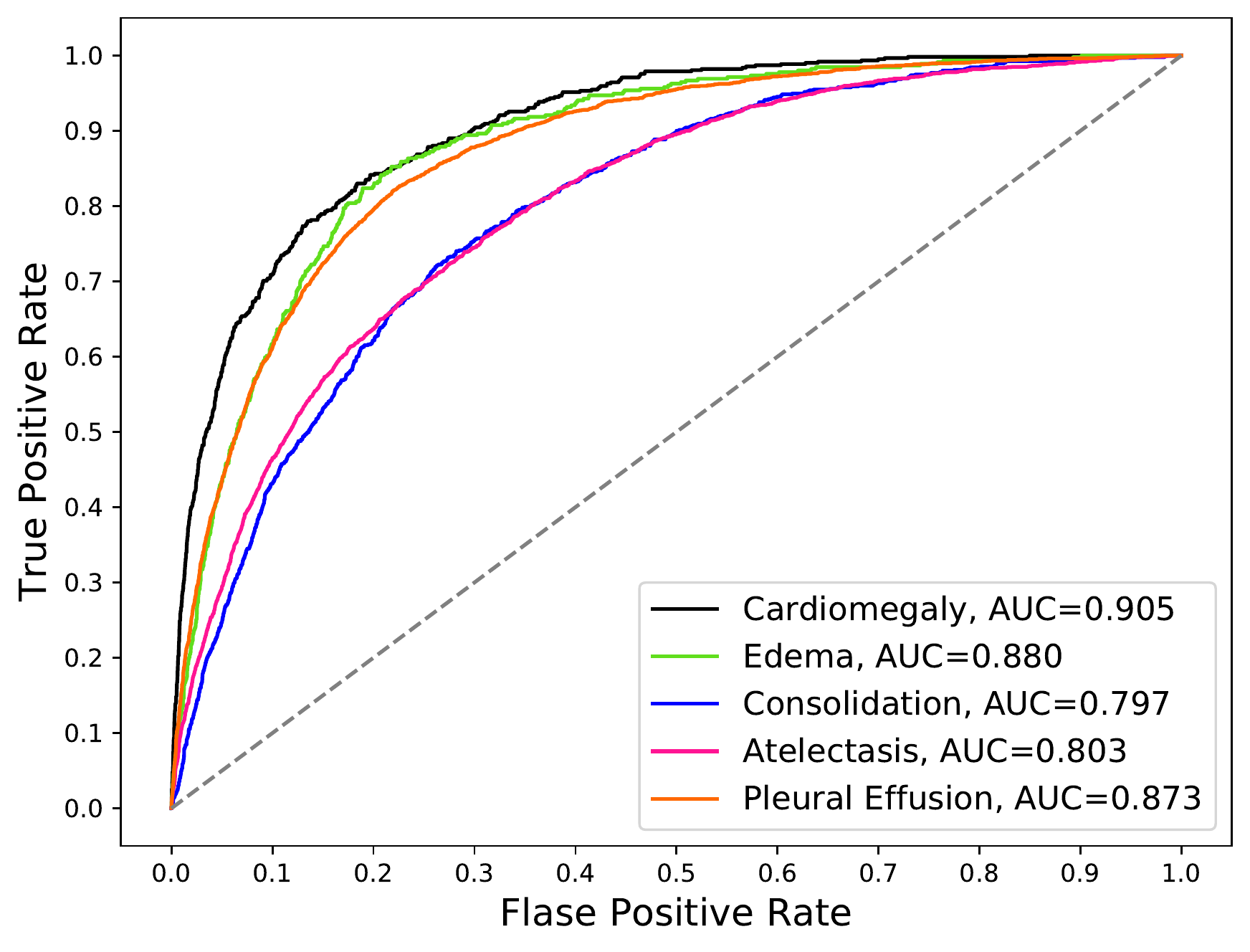}
\centering 
\captionsetup{justification=centering}
\caption{Train: ChestX-ray14 \\ Test:Internal}
\label{DenseNet2_NIH}
\end{subfigure}
\begin{subfigure}{0.32\textwidth}
\centering
\includegraphics[width=\linewidth]{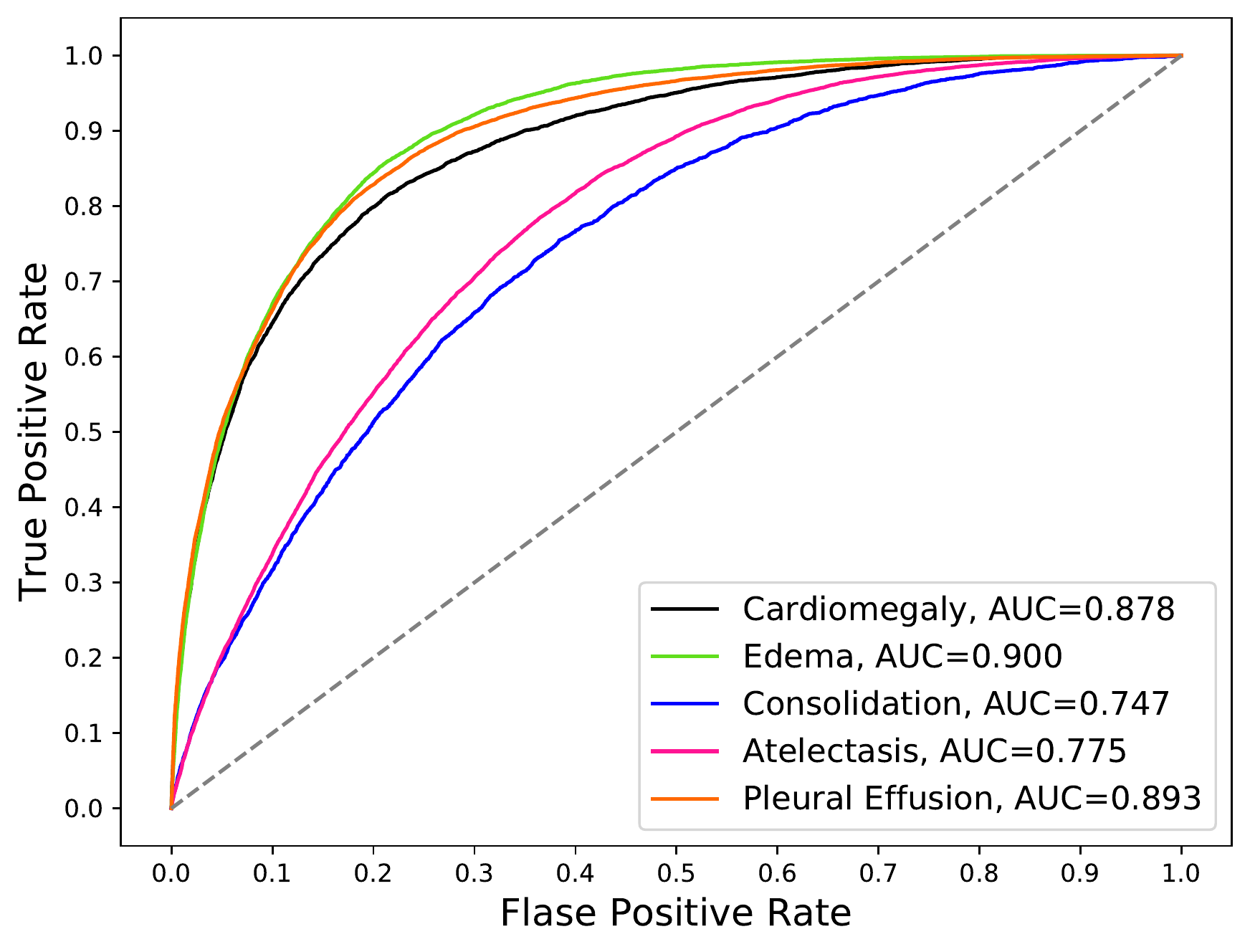}
\centering 
\captionsetup{justification=centering}
\caption{Train: CheXpert-ChestX-ray14 \\ Test: Internal}
\label{DenseNet3_chex_NIH}
\end{subfigure}

\begin{subfigure}{0.32\textwidth}
\vspace{2.5mm}
\centering
\includegraphics[width=\linewidth]{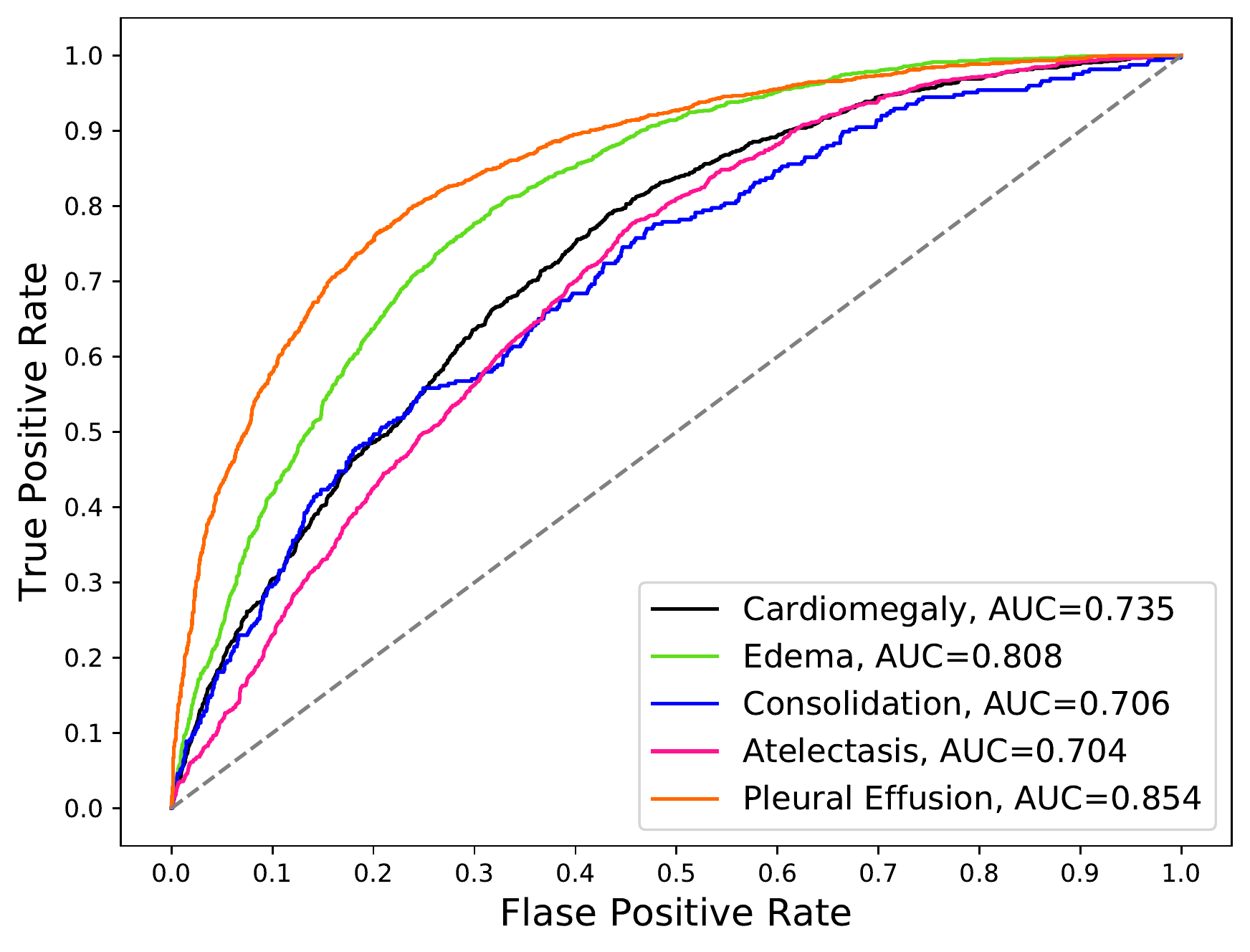}
\centering 
\captionsetup{justification=centering}
\caption{Train: CheXpert \\Test: External}
\label{DenseNet1_mimic}
\end{subfigure}
\begin{subfigure}{0.32\textwidth}
\vspace{2.5mm}
\centering
\includegraphics[width=\linewidth]{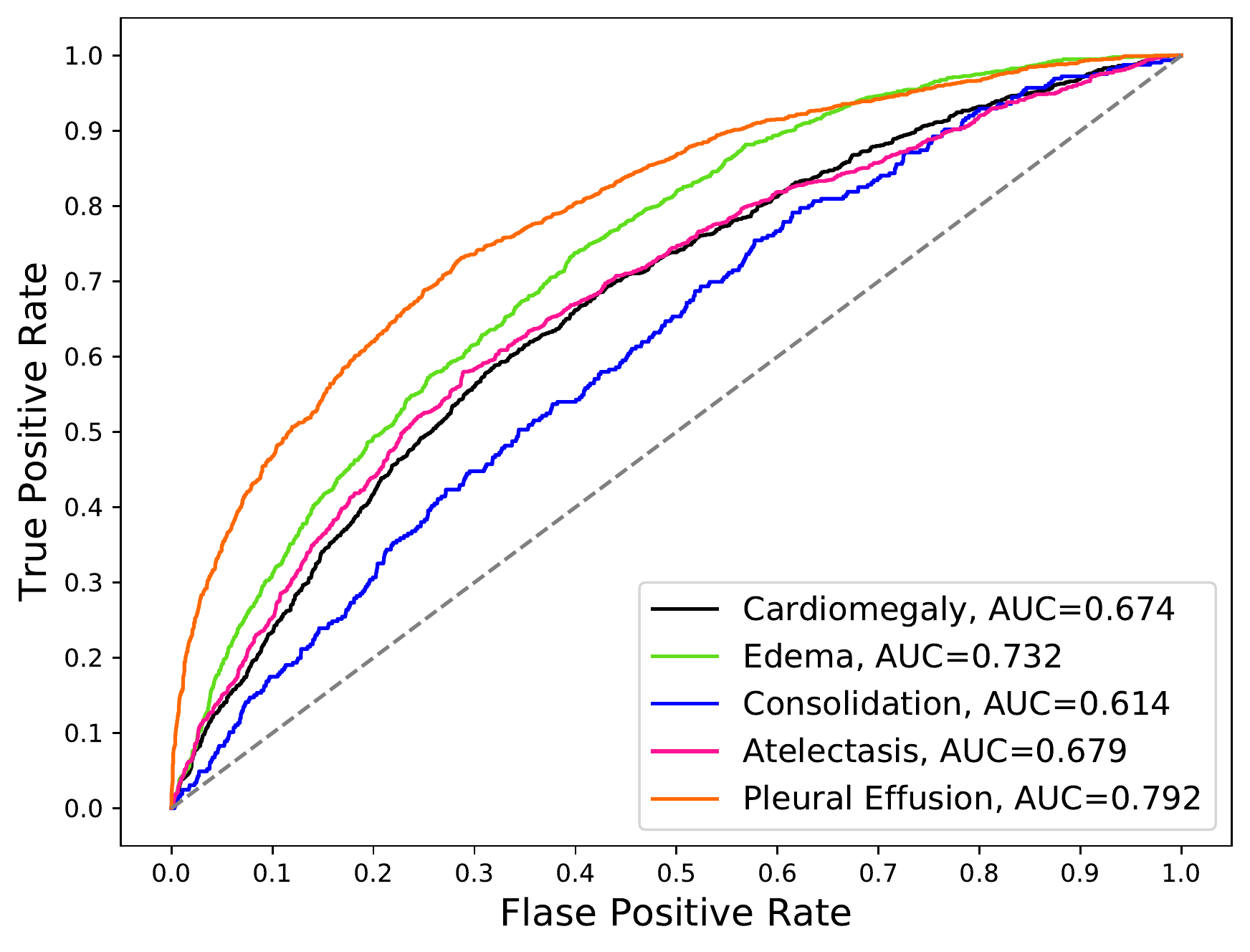}
\centering 
\captionsetup{justification=centering}
\caption{Train: ChestX-ray14 \\Test: External}
\label{DenseNet2_mimic}
\end{subfigure}
\begin{subfigure}{0.32\textwidth}
\vspace{2.5mm}
\centering
\includegraphics[width=\linewidth]{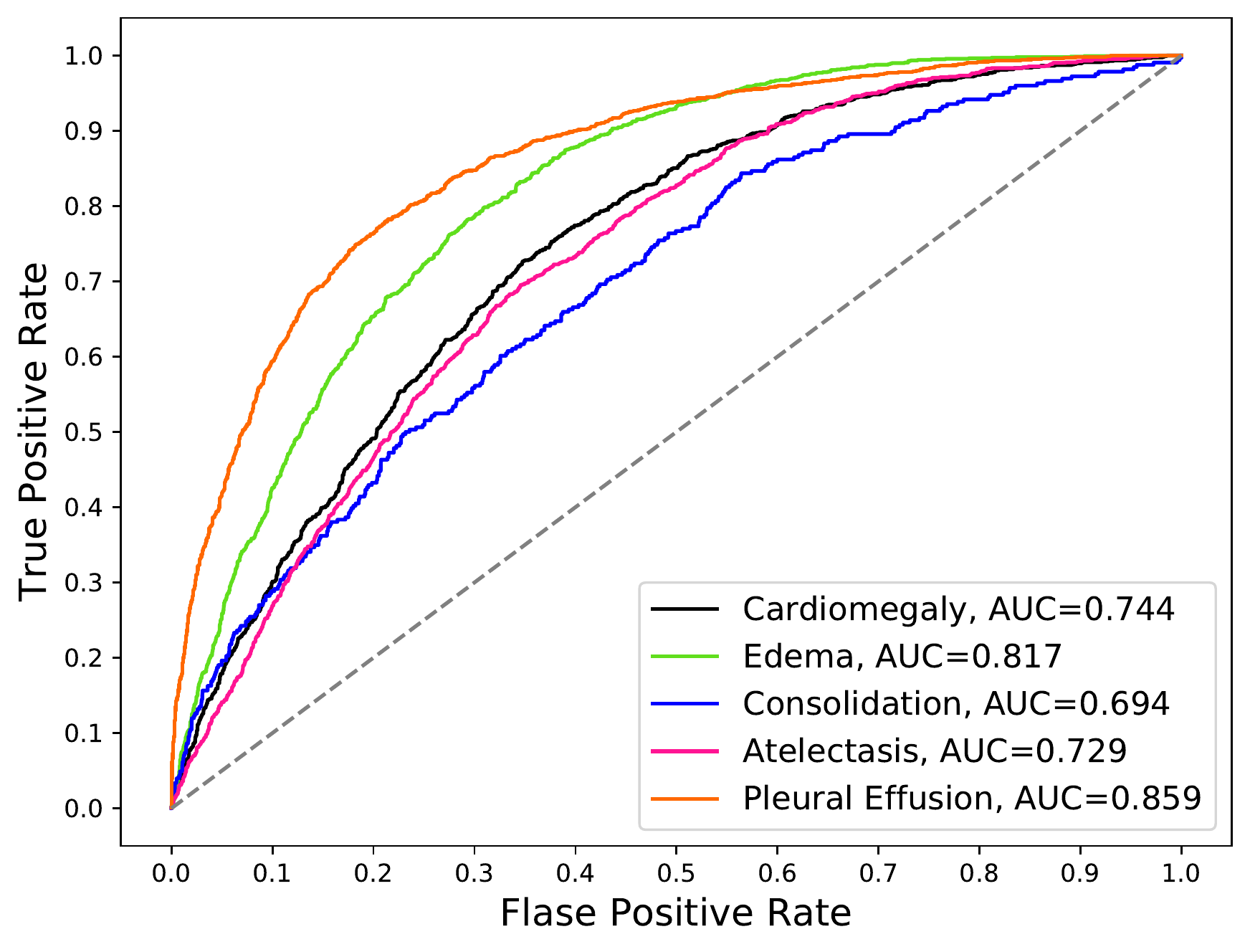}
\centering 
\captionsetup{justification=centering}
\caption{Train: CheXpert-ChestX-ray14 \\Test: External}
\label{DenseNet3_mimic}
\end{subfigure}
\caption{\label{fg:densenet121_curves} ROC curves for DenseNet121 for all evaluation configurations}
\end{figure}

\begin{figure}[H]
\vspace{2.5mm}
\begin{subfigure}{0.32\textwidth}
\centering
\includegraphics[width=\linewidth]{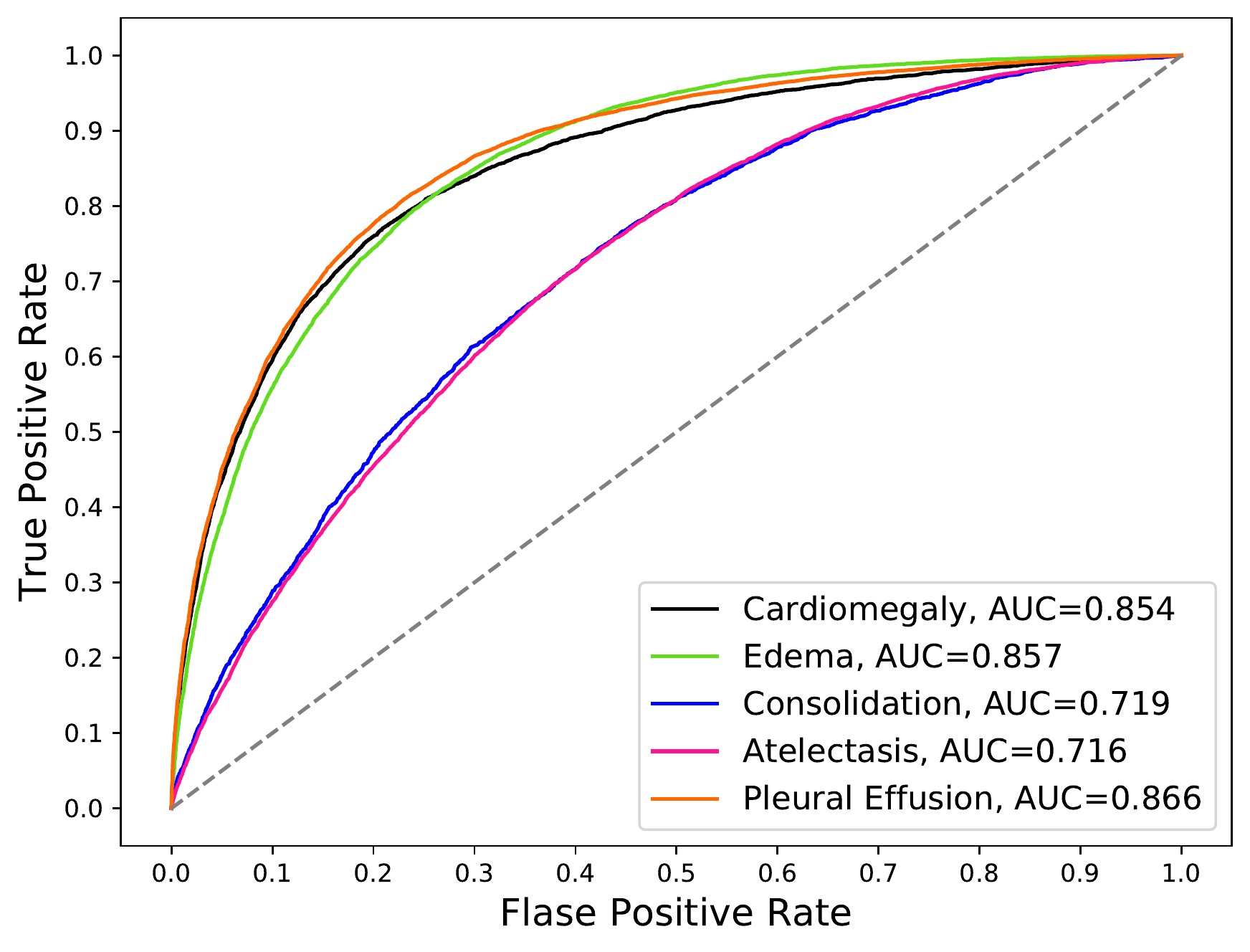}
\centering 
\captionsetup{justification=centering}
\caption{Train: CheXpert \\Test: Internal}
\label{inception1_chex}
\end{subfigure}
\begin{subfigure}{0.32\textwidth}
\centering
\includegraphics[width=\linewidth]{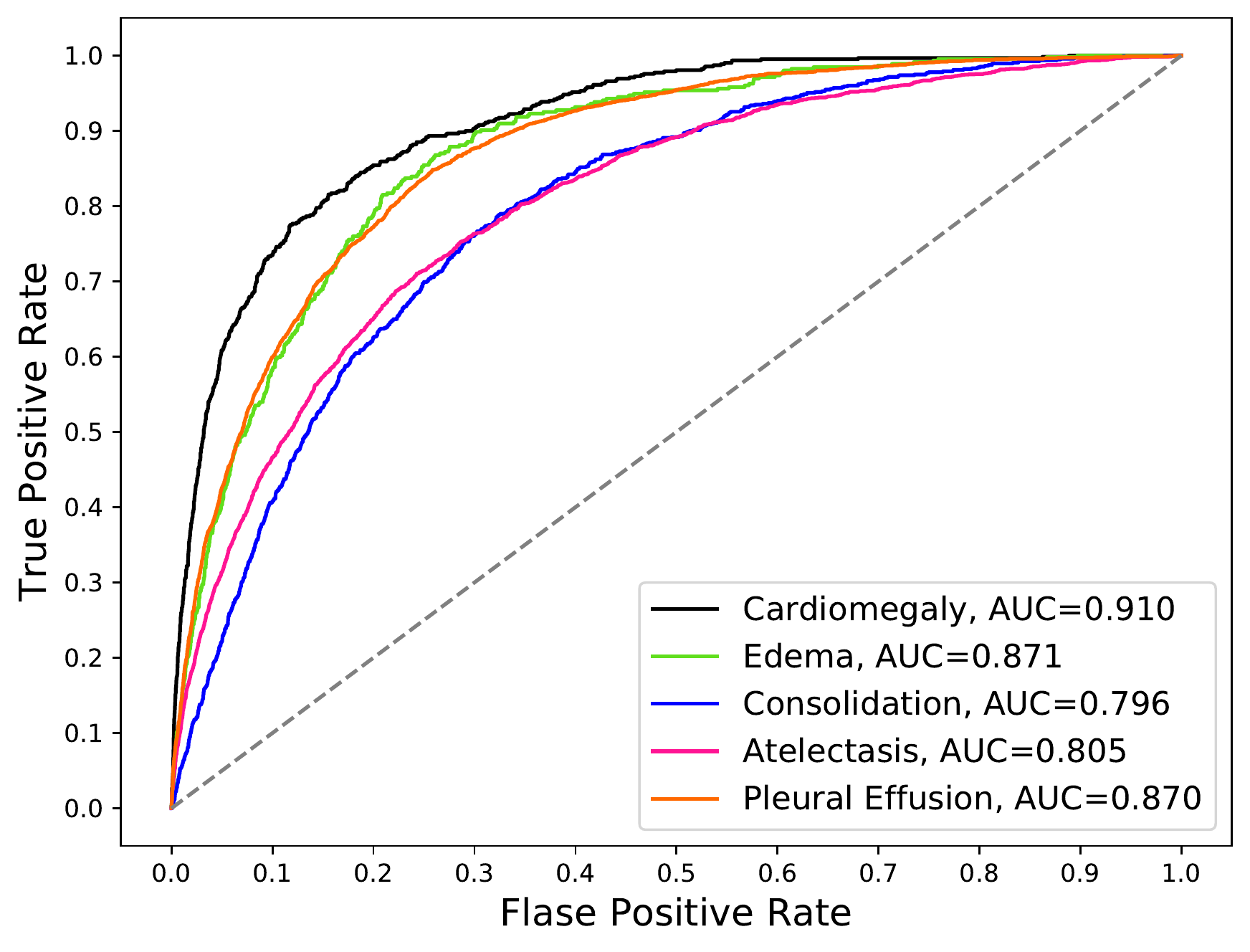}
\centering 
\captionsetup{justification=centering}
\caption{Train: ChestX-ray14 \\Test: Internal}
\label{inception2_NIH}
\end{subfigure}
\begin{subfigure}{0.32\textwidth}
\centering
\includegraphics[width=\linewidth]{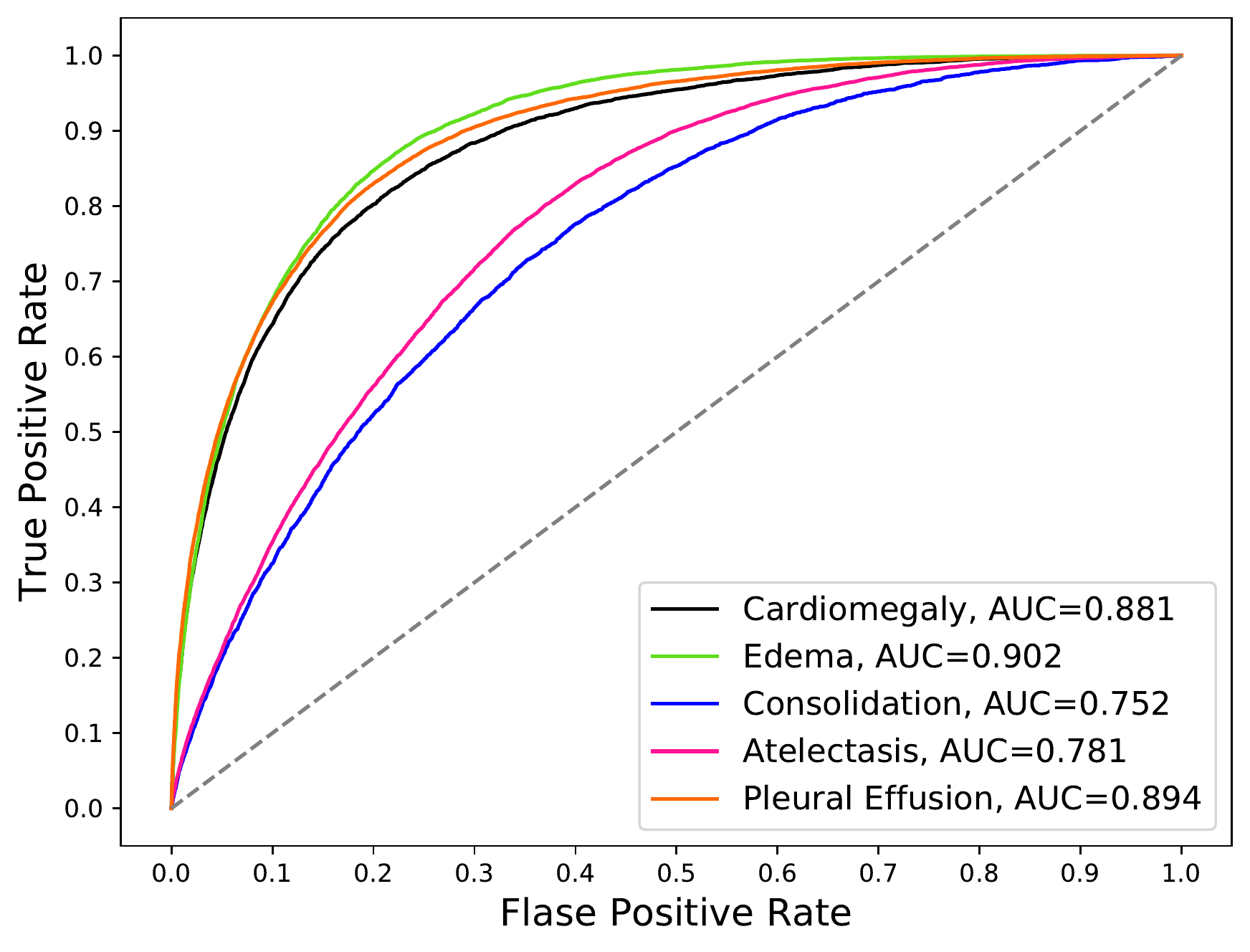}
\centering 
\captionsetup{justification=centering}
\caption{Train: CheXpert-ChestX-ray14 \\Test: Internal}
\label{inception3_chex_NIH}
\end{subfigure}

\begin{subfigure}{0.32\textwidth}
\vspace{2.5mm}
\centering
\includegraphics[width=\linewidth]{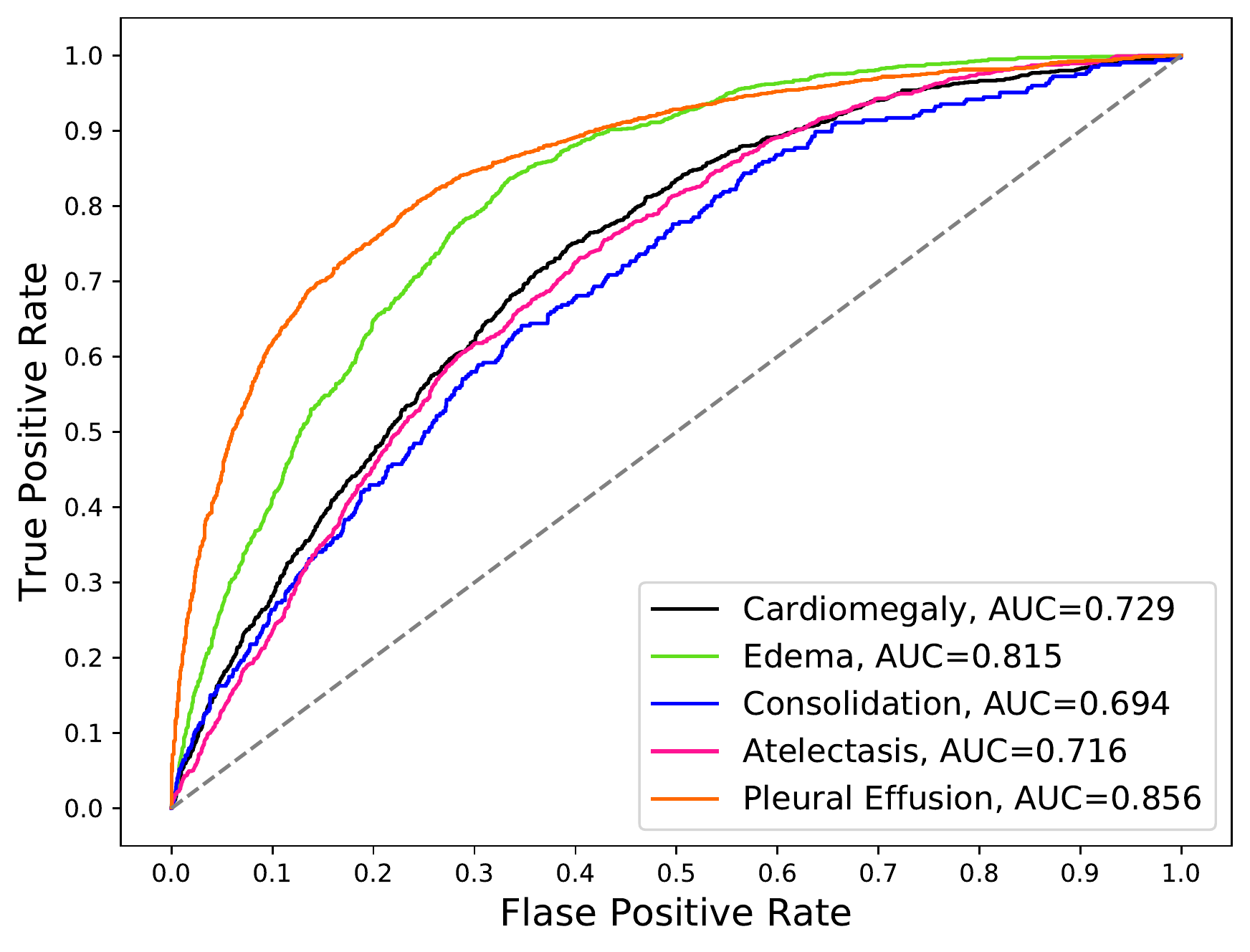}
\centering 
\captionsetup{justification=centering}
\caption{Train: CheXpert \\Test: External}
\label{inception1_mimic}
\end{subfigure}
\begin{subfigure}{0.32\textwidth}
\vspace{2.5mm}
\centering
\includegraphics[width=\linewidth]{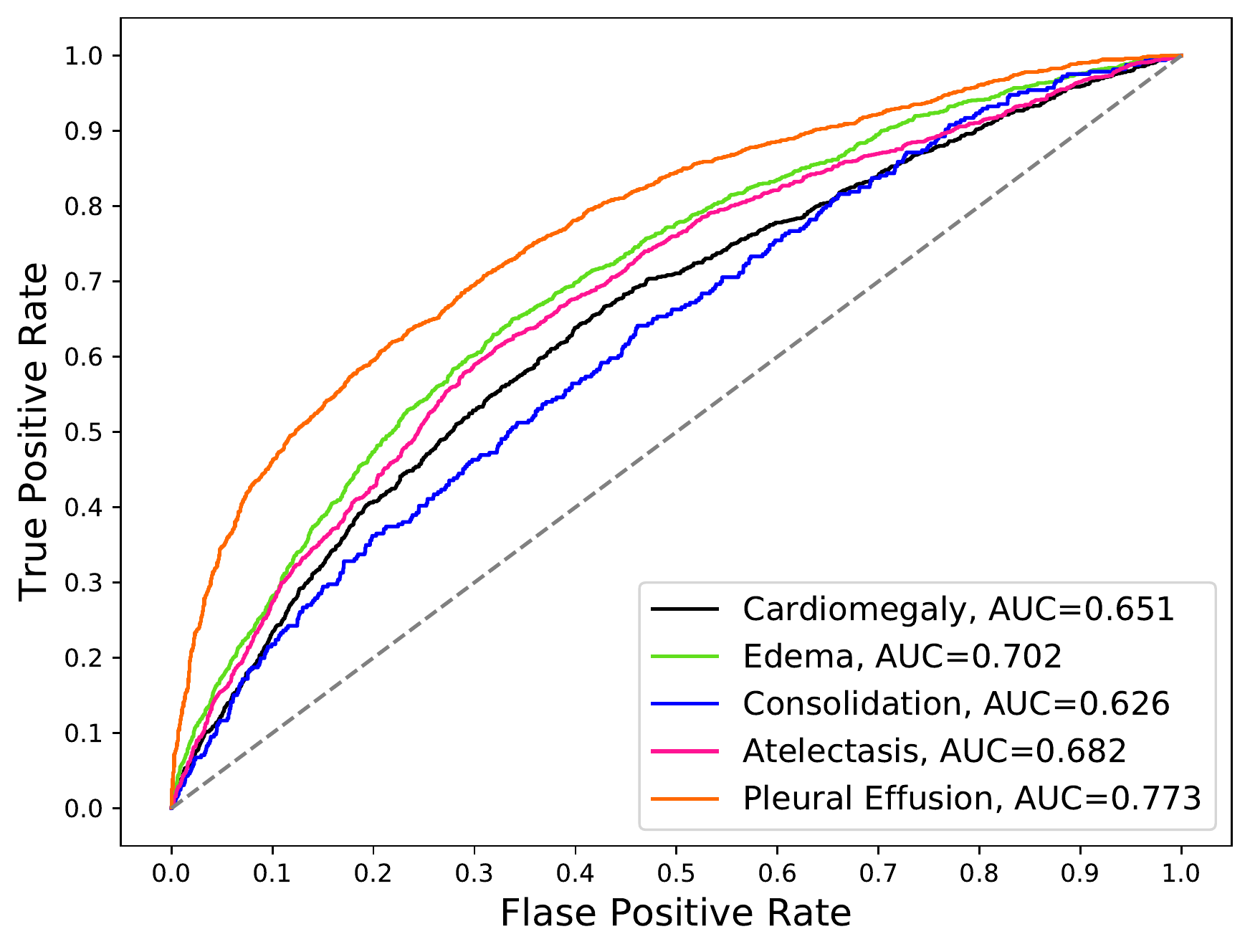}
\centering 
\captionsetup{justification=centering}
\caption{Train: ChestX-ray14 \\ Test: External}
\label{inception2_mimic}
\end{subfigure}
\begin{subfigure}{0.32\textwidth}
\vspace{2.5mm}
\centering
\includegraphics[width=\linewidth]{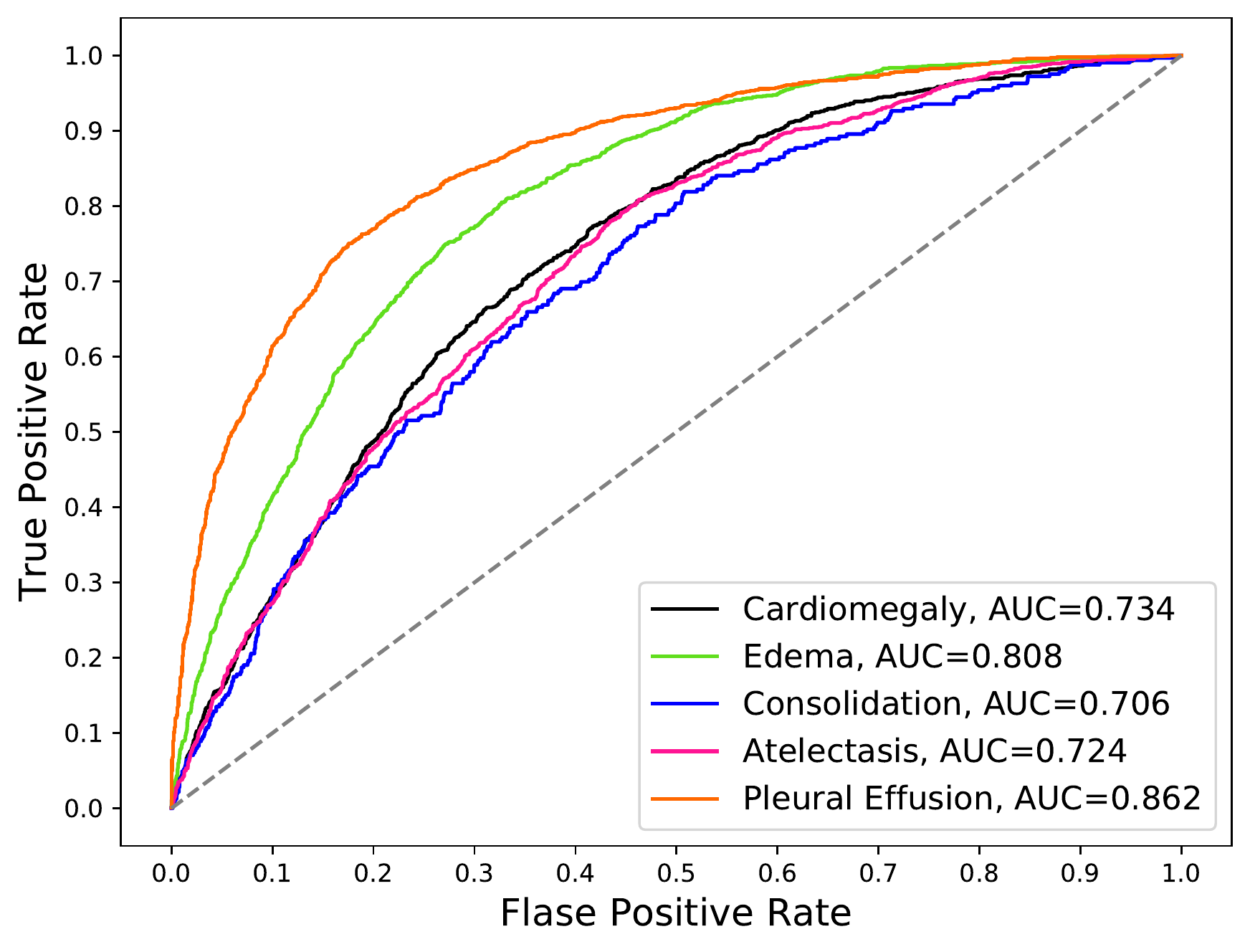}
\centering 
\captionsetup{justification=centering}
\caption{Train: CheXpert-ChestX-ray14 \\ Test: External}
\label{inception3_mimic}
\end{subfigure}

\caption{\label{fg:inceptionresnetv2_curves} ROC curves for InceptionResNetV2 for all evaluation configurations}
\end{figure}

Increasing the difference in the prevalence of all the $5$ diseases changes model performance with improved AUCs on the external sets when compared to corresponding values of other configurations except for two cases. First, DenseNet121 trained at CheXpert to detect Consolidation has better external test AUC $(0.706)$ compared to joint CheXpert-ChestX-ray14 (AUC $= 0.694$). Second, InceptionResNetV2 trained at CheXpert to detect Edema has better external test AUC $(0.815)$ compared to joint CheXpert-ChestX-ray14 (AUC $= 0.808$).

ROC curves for all evaluation configurations are displayed in Figures \ref{fg:densenet121_curves} and \ref{fg:inceptionresnetv2_curves} for DenseNet121 and InceptionResNetV2 respectively. It is evident that ROC curves have larger areas under the curve for internal test sets for all configurations for both models than corresponding curves for external test sets. For the joint training set (CheXpert-ChestX-ray14), curves have larger areas under the curve for the external test than the other two configurations for both models. This trend indicates better generalization capacity of the trained model under joint training configuration.%Curves are slightly worsened for the internal test under this joint training condition.

\subsection*{Radiologist Evaluation}
We observed that the models tend to perform very differently for different labels. For example, the performance of both models is quite poor for Cardiomegaly under all evaluation configurations. We consulted a board-certified radiologist to further investigate this trend. The radiologist reviewed randomly chosen MIMIC-CXR-JPG images (external test set) of Cardiomegaly label that were incorrectly labeled by trained models under configuration $3$, i.e., trained over joint CheXpert and ChestX-ray14 datasets. Table \ref{tb:sample_cases} shows a few samples of such images with their groundtruth label and predicted probability of that label for each model.

% \begin{table}[H]
% \centering
% \caption{Incorrectly classified radiographic images of borderline Cardiomegaly pathology; Groundtruth label for each image as well as probabilities of that label predicted by both models are included. The maximum diameter of the heart and the thoracic diameter were measured by the radiologist (shown by red lines) to estimate cardiothoracic ratio to diagnose Cardiomegaly. All images were classified as \textbf{Borderline Cardiomegaly} by the radiologist.}
% \label{tb:sample_cases}
% \begin{tabular}{|c|l|c|c|}
% \hline
% \textbf{Sample Images} & \textbf{Groundtruth Label} & \textbf{InceptionResNetV2} & \textbf{DenseNet121} \\ \hline
% \begin{minipage}{.13\textwidth}
%       \includegraphics[width=\linewidth]{amia_template/pics/Cardiomegaly/a.jpg}
%     \end{minipage} & No Cardiomegaly & $0.496$ & $0.696$ 
%      \\ \hline
%  \begin{minipage}{.13\textwidth}
%       \includegraphics[width=\linewidth]{amia_template/pics/Cardiomegaly/b.jpg}
%     \end{minipage}& Cardiomegaly & $0.605$ & $0.477$  \\ \hline
%  \begin{minipage}{.13\textwidth}
%       \includegraphics[width=\linewidth]{amia_template/pics/Cardiomegaly/c.jpg}
%     \end{minipage}& No Cardiomegaly & $0.472$ & $0.553$  \\ \hline
%  \begin{minipage}{.13\textwidth}
%       \includegraphics[width=\linewidth]{amia_template/pics/Cardiomegaly/d.jpg}
%     \end{minipage}& No Cardiomegaly & $0.565$ & $0.419$ \\ \hline
% \end{tabular}
% \end{table}

Radiologists rely on the cardiothoracic ratio to diagnose Cardiomegaly. It is the ratio of maximal horizontal cardiac diameter and maximal horizontal thoracic diameter ($inner\;edge\;of\;ribs / edge \;of\;pleura$) and is measured on a Posteroanterior (PA) chest X-ray. A normal measurement should be less than $0.5$. The first and third images shown in Table \ref{tb:sample_cases} were incorrectly predicted to have Cardiomegaly by DenseNet121. However, InceptionResNetV2 was able to correctly predict the absence of Cardiomegaly for these images. Also, the DenseNet121 model predicted incorrectly that the second image shown in Table \ref{tb:sample_cases} does not have Cardiomegaly. InceptionResNetV2 correctly predicted Cardiomegaly for this image. On the other hand, InceptionResNetV2 incorrectly predicted the fourth image shown in Table \ref{tb:sample_cases} to have Cardiomegaly. Radiologist assessed that these are all borderline Cardiomegaly cases based on the cardiothoracic ratio. The AI models performed poorly for borderline cases. Another observation is that the pacemaker is closely tied with Cardiomegaly for both models. The presence of this foreign object may bias the decision of the DCNN models.  

\begin{table}[H]
\centering
\medskip
\caption{Incorrectly classified radiographic images of borderline Cardiomegaly pathology; Groundtruth label for each image as well as probabilities of that label predicted by both models are included. The maximum diameter of the heart and the thoracic diameter were measured by the radiologist (shown by red lines) to estimate cardiothoracic ratio to diagnose Cardiomegaly. All images were classified as \textbf{Borderline Cardiomegaly} by the radiologist.}
\label{tb:sample_cases}
\begin{tabular}{|l|c|c|c|c|}
\hline
 \multirow{4.5}{*}{\textbf{Sample Images}}                           & $1$               & $2$            & $3$               & $4$               \\ 
     & \begin{minipage}{.17\textwidth}
      \includegraphics[width=\linewidth]{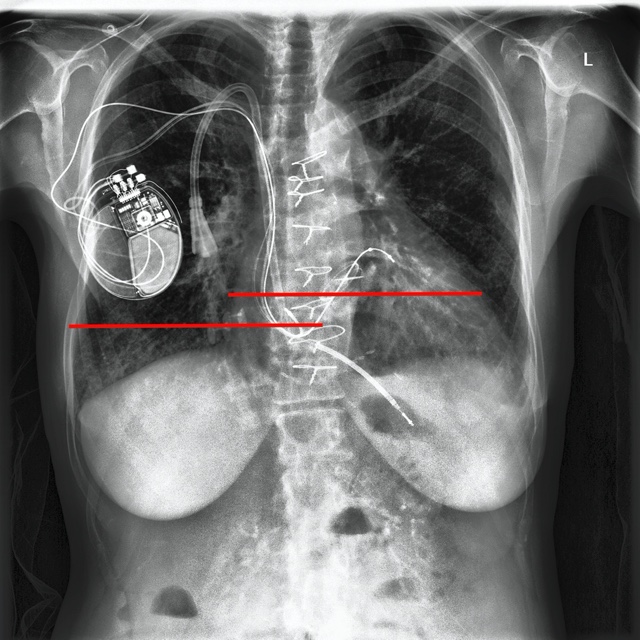}
    \end{minipage}               & \begin{minipage}{.17\textwidth}
      \includegraphics[width=\linewidth]{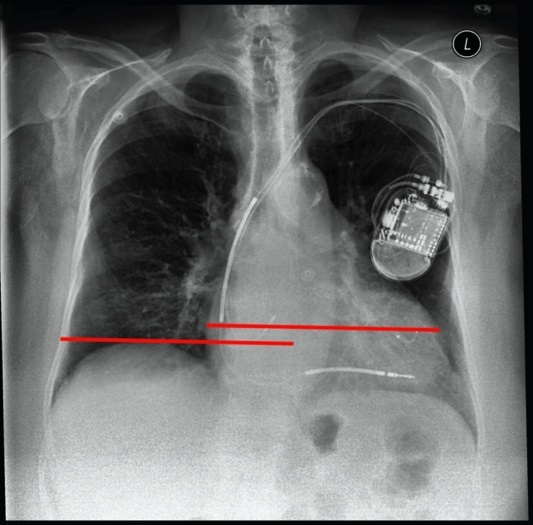}
    \end{minipage}           & \begin{minipage}{.17\textwidth}
      \includegraphics[width=\linewidth]{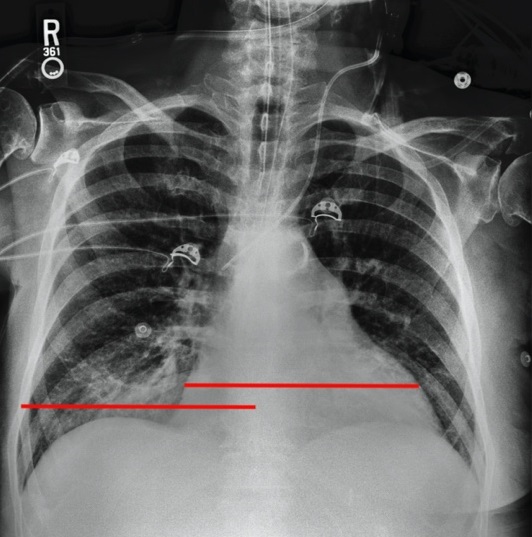}
    \end{minipage}               & \begin{minipage}{.17\textwidth}
      \includegraphics[width=\linewidth]{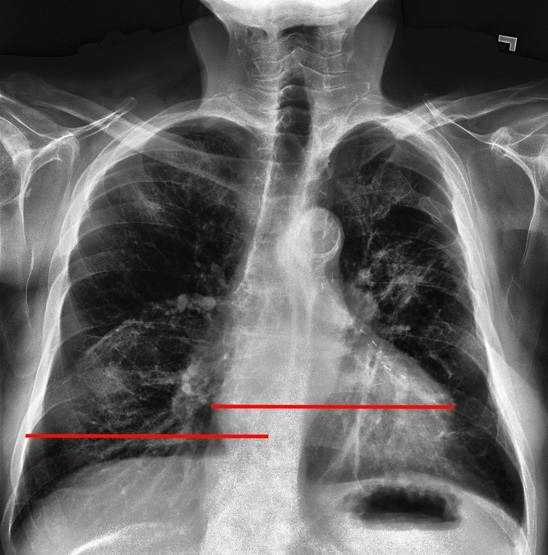}
    \end{minipage}               \\ \hline
\textbf{Groundtruth Label} & No Cardiomegaly & Cardiomegaly & No Cardiomegaly & No Cardiomegaly \\ \hline
\textbf{InceptionResNetV2} & 0.496           & 0.605        & 0.472           & 0.565           \\ \hline
\textbf{DenseNet121}       & 0.696           & 0.477        & 0.553           & 0.419 \\ \hline        
\end{tabular}
\end{table}

\section*{Conclusion}
In this paper, we thoroughly studied the robustness and generalization capabilities of complex deep learning classification models. We used three publicly available chest X-ray datasets (CheXpert, ChestX-ray14, MIMIC-CXR-JPG) and experimented with two different DCNN models (InceptionResNetV2, DenseNet121). Our experiments indicate that these models have limited generalization capacity when tested over images outside of the training dataset, i.e., external test set. For all class labels under every evaluation configuration, the performance of each model is better for the internal test set than it is for the external test. We also worked on improving generalization capabilities of these models. Our technique relies on improving the quality of the training data by combining images from different datasets, thus increasing the data variation the models are exposed to during the training phase. This technique proved effective as performance of models was significantly improved for external test sets. In this case, even incorrectly predicted labels tend to be borderline cases of their corresponding pathologies. Interestingly, the performance of each model for internal test sets remains approximately the same level. Therefore, we can conclude that generalization of deep learning classification models to a larger variety of items is heavily dependent on the quality and heterogeneity of the training dataset. Exposing the model to multiple datasets with wide variation during the training phase is an effective technique for improvement in the generalization capabilities of the trained model.

\makeatletter
\renewcommand{\@biblabel}[1]{\hfill #1.}

\makeatother
\let\oldbibliography\thebibliography
\renewcommand{\thebibliography}[1]{%
  \oldbibliography{#1}%
  \setlength{\itemsep}{-0.1em}%
}
\begin{center}
\bibliography{bibliography}
\end{center}

\end{document}